%% file: main.tex
\documentclass[10pt, twocolumn, letterpaper]{article}

\usepackage[pagenumbers]{cvpr} % To force page numbers, e.g. for an arXiv version

\input{preamble}

\definecolor{cvprblue}{rgb}{0.21,0.49,0.74}
\usepackage[pagebackref, breaklinks, colorlinks, allcolors=cvprblue]{hyperref}

\def\paperID{14310} % *** Enter the Paper ID here
\def\confName{CVPR}
\def\confYear{2026}

\title{\includegraphics[height=0.8em]{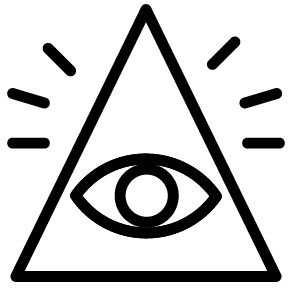}\method: An All-Seeing Diffusion Model for 3D and 4D View Synthesis}

\author{
Xiang Fan$^{1,2}$ \and \hspace{-10pt}
Sharath Girish$^{2}$ \and  \hspace{-10pt}
Vivek Ramanujan$^{1}$ \and  \hspace{-10pt}
Chaoyang Wang$^{2}$ \and  \hspace{-10pt}
Ashkan Mirzaei$^{2}$ \and  \hspace{-10pt}
Petr Sushko$^{1}$ \and  \hspace{-10pt}
Aliaksandr Siarohin$^{2}$ \and  \hspace{-10pt}
Sergey Tulyakov$^{2}$ \and  \hspace{-10pt}
Ranjay Krishna$^{1}$\and \\[-5pt]
$^{1}$University of Washington \quad $^{2}$Snap Inc. \\
Project page: \href{https://snap-research.github.io/OmniView/}{https://snap-research.github.io/OmniView}
}

\begin{document}
    % \maketitle
    \input{figures/fig_teaser}
    \input{sec/0_abstract}
    \input{sec/1_intro}

    \input{sec/2_related}
    \input{sec/3_method}

    \input{sec/4_experiments}
    \input{sec/5_conclusion}
    \newpage
    { \small \bibliographystyle{ieeenat_fullname} \bibliography{main} }

    \input{sec/X_suppl}
\end{document}

%% file: preamble.tex
\usepackage{algorithm}
\usepackage{algpseudocode}
\usepackage{amsfonts}
\usepackage{amsmath}
\usepackage{booktabs}
\usepackage{enumitem}
\usepackage{graphicx}
\usepackage{hhline}
\usepackage{mathtools}
\usepackage{multirow}
\usepackage{nicefrac}
\usepackage{pifont}
\usepackage{soul}
\usepackage{tabularx}
\usepackage{booktabs}
\usepackage{array}
\usepackage{makecell}
\usepackage{url}
\usepackage{wrapfig}
\newcommand{\method}{\textsc{OmniView}\xspace}

\newcommand{\thetaxyt}{\boldsymbol{\theta}_{xyt}}

\renewcommand{\paragraph}[1]{\vspace{.5em}\noindent\textbf{#1.}}

\newcolumntype{Y}{>{\centering\arraybackslash}X}
\newcolumntype{Z}{>{\hspace{0.4em}\centering\arraybackslash}X<{\hspace{0.4em}}}

%% file: figures/fig_teaser.tex
\twocolumn[{%
\maketitle
\vspace{-0.6em}

\begin{center}
  \includegraphics[width=\linewidth]{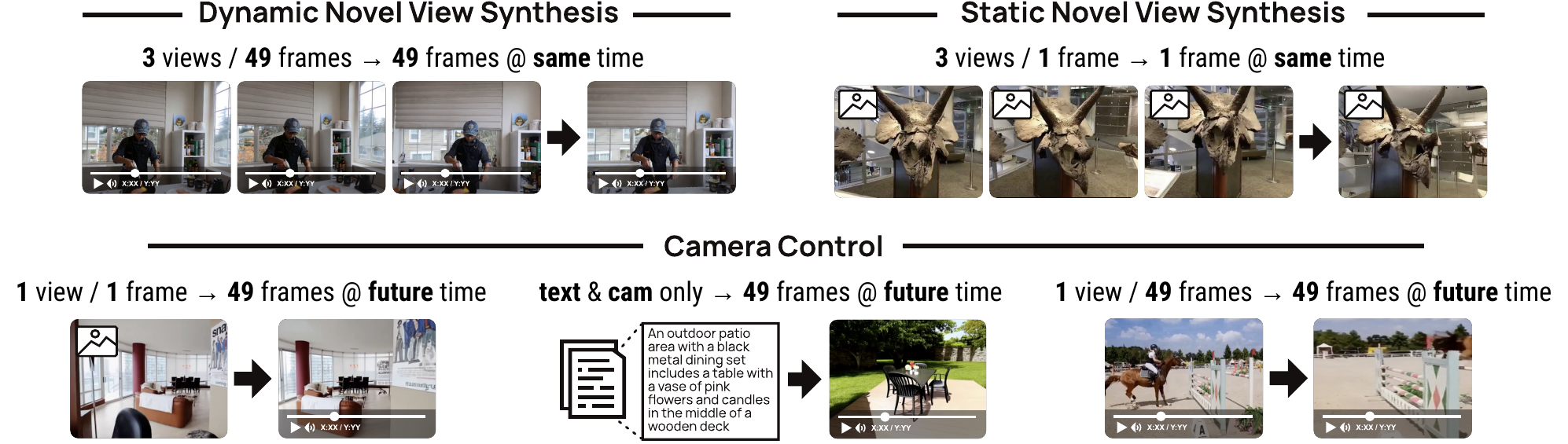}
  \captionof{figure}{\textbf{\method: } Given one or more images or videos, captured at potentially different viewpoints and times, \method generates 4D-consistent videos that can be rendered from novel camera trajectories, viewpoints, times, and durations. \method unifies many existing video generation paradigms, enabling tasks such as novel view synthesis (NVS), text-to-video (T2V) camera control, and multi-view conditioning in one model.}
  \label{fig:teaser}
\end{center}
\vspace{1em}
}]

%% file: sec/0_abstract.tex
\begin{abstract}
% Achieving general-purpose spatiotemporal consistency in video generation remains a key challenge. 
Prior approaches injecting camera control into diffusion models have focused on specific subsets of 4D consistency tasks: novel view synthesis, text-to-video with camera control, image-to-video, amongst others.
Therefore, these fragmented approaches are trained on disjoint slices of available 3D/4D data.
We introduce \method, a unified framework that generalizes across a wide range of 4D consistency tasks. Our method separately represents space, time, and view conditions, enabling flexible combinations of these inputs. For example, \method can synthesize novel views from static, dynamic, and multiview inputs, extrapolate trajectories forward and backward in time, and create videos from text or image prompts with full camera control.
\method is competitive with task-specific models across diverse benchmarks and metrics, improving image quality scores among camera-conditioned diffusion models by up to 33\% in multiview NVS LLFF dataset, 60\% in dynamic NVS Neural 3D Video benchmark, 20\% in static camera control on RE-10K, and reducing camera trajectory errors by 4x in text-conditioned video generation. With strong generalizability in one model, \method demonstrates the feasibility of a generalist 4D video model. 

% Our approach introduces a single unified generative model which is flexible in the number and type of input conditions such as text/images/videos. We propose an architecture with a space-time RoPE mechanism in DiTs to allow jointly combining camera inputs and video timesteps. We introduce a training mechanism for combining a wide variety of existing 3D datasets towards tackling the different tasks. We show applications in a variety of view synthesis tasks achieving competitive performance or even outperforming task-specific models. We show, for the first time, that video generative models can also target the multi-view video reconstruction task achieving consistent generations without any explicit 3D conditions such as depth or point clouds.  
\end{abstract}

%% file: sec/1_intro.tex
\section{Introduction}
\label{sec:intro}

When trained on raw internet-scale video, video diffusion models internalize strong 3D priors~\cite{snapvideo,sv3d,li2024director3d}.
They can synthesize long, coherent camera motions and maintain scene layout without any explicit geometry~\cite{genie3}. 
Yet, by default, they offer little explicit 3D control. 
To deploy them in applications such as virtual/augmented reality, film production, robotics, or autonomous driving, they need to be controllable. 
Many applications would become possible if we could control how the generator's camera moves in space and in time. 
Naturally, this motivation has driven multiple sub-fields of computer vision to develop specialized solutions. 
Camera redirection now exists for multi-view static~\cite{gao2024cat3d,zhou2025stable} and dynamic Novel View Synthesis (NVS)~\cite{syncammaster,wang20254real}, text-to-video (T2V)~\cite{cameractrl,dav24} and image-to-video (I2V)~\cite{camco} with camera control, and even video-to-video (V2V)~\cite{bai2025recammastercameracontrolledgenerativerendering}.

However, existing approaches are fragmented along task, architecture, and data. 
Methods designed for multi-view static NVS focus on reconstructing a single scene from sparse views~\cite{zhou2025stable}, achieving strong 3D consistency but only at a fixed timestamp, and thus cannot handle dynamic videos. Camera-control T2V/I2V models convert text or a single image plus a camera trajectory and generate a moving video, but their architectures cannot ingest full input videos~\cite{cameractrl}. 
V2V redirection models accept a source video and re-render it from a new camera path at matched timestamps, yet they typically cannot operate in the multi-view setting~\cite{bai2025recammastercameracontrolledgenerativerendering}. 
Some works rely on explicit geometric representations (depth maps, point clouds, or other 3D/4D fields) for consistency~\cite{yu2025trajectorycrafter,ren2025gen3c,yu2024viewcrafter,wu2024reconfusion}, rather than exploiting the implicit 3D priors already present in video models.

Because each family of methods is tailored to a narrow I/O configuration, they are trained on disjoint slices of available 3D/4D data, leaving most geometric supervision unused. 
We argue that a single, flexible framework that can express all these tasks, and train on heterogeneous datasets, should both improve generalization across 3D tasks and substantially reduce deployment overhead.

Our approach, \method, instantiates such a unified framework as a single video generative model for diverse view-synthesis tasks. We model each image $\mathbf{I}$ as a sample from a 4D world, parameterized by a camera pose $\mathbf{p}$ and a timestamp $t$. Under this view, static multi-view NVS corresponds to varying $\mathbf{p}_{1:N}$ and a target pose $\mathbf{p}^{t}$ while keeping $t$ fixed; I2V with camera control corresponds to predicting frames at future times $t_{1:N}$ along target poses $\mathbf{p}^{t}_{1:N}$ given an input $(\mathbf{I}_0, \mathbf{p}_0, t_0)$; and V2V camera redirection corresponds to re-rendering an input video from new poses $\mathbf{p}^{t}_{1:N}$ at the same timestamps $t_{1:N}$ as the source.

To realize this unified 4D formulation, we adopt a Diffusion Transformer (DiT)~\cite{dit} backbone that naturally handles a variable number of input tokens. We tokenize each frame into a set of video tokens, concatenate tokens from all available inputs (images, views, or frames), and condition generation on this sequence. DiTs already support temporal reasoning via spatio-temporal Rotary Positional Embeddings (3D RoPE), which encode $(x, y, t)$ for each token. Prior works typically injects camera information by either encoding poses with a pose encoder or by mapping them to Plücker ray embeddings and then applying 3D RoPE~\cite{su2024roformer} to both video and camera-conditioned tokens. This entangles camera pose $\mathbf{p}\mathbf{p}$ and time $t$ in a single positional embedding space, making it difficult for the model to learn 3D structure independently of temporal dynamics and often leading to overfitting to seen trajectories.

In contrast, \method explicitly disentangles space and time. We represent each token's camera pose as Plücker rays and apply \textit{spatial} 2D RoPE only to these Plücker features, then concatenate them channel-wise with the corresponding video token. Time is encoded separately via temporal RoPE on the video tokens. This design cleanly separates camera geometry from temporal evolution, while still allowing the DiT to jointly attend over all tokens. Combined with the variable-token design, this lets \method flexibly ingest arbitrary combinations of frames, views, and timestamps and thereby support a wide range of 4D inputs under a single architecture. We then devise a joint training strategy that mixes heterogeneous 3D datasets, each corresponding to different task configurations (multi-view static, dynamic, T2V/I2V with camera control, V2V redirection), so that the model learns shared geometric priors across them.

We extensively evaluate \method on static and dynamic NVS benchmarks with monocular and multi-view inputs, as well as camera-control tasks conditioned on text and images. \method consistently matches 
or outperforms specialized baselines, producing high-fidelity, 3D-consistent videos.
With up to 33\% increase in SSIM scores in multiview NVS LLFF dataset, 60\% in dynamic NVS Neural 3D Video benchmark, 20\% in static camera control on RE-10K, 4x reduction in camera errors in text-conditioned video generation, \method demonstrates strong 3D consistency and fidelity across tasks, and generalization to input configurations not seen during training. Ablations validate the benefit of our RoPE design.

%% file: sec/2_related.tex
\section{Related Work}
\label{sec:related}

% \subsection{Camera control}
% \subsection{Single/multiview}
% \subsection{Video-to-video}
% \subsection{Implicit vs explicit?}

\paragraph{Camera-Controllable Video Generation}
The emergence of powerful text-to-video diffusion models~\cite{blattmann2023stable, videocrafter2, cogvideox, snapvideo, hunyuanvideo} has fueled extensive research on conditioning generated videos with additional controls, such as camera parameters~\cite{direct_a_video, vd3d, cami2v, zheng2025vidcraft3, liu2023zero1to3, shi2023zero123plus, vanhoorick2024gcd, animatediff, MotionCtrl, watson2024controlling}. Early camera-control approaches integrate extrinsic camera information as part of the diffusion model’s conditioning, either through tailored encoders or numeric input channels. Models like MotionCtrl~\cite{MotionCtrl} and CameraCtrl~\cite{cameractrl} encode camera poses or trajectories to enable user-directed viewpoint changes throughout video generation, but often require specific paired training data and show limited generalization when camera motions deviate from training regimes. Other strategies bypass model retraining by employing 3D geometric cues, for example warping frames using estimated depth to match new camera placements and feeding these as priors during the denoising process~\cite{NVS_Solver, camtrol}, though these methods face a trade-off between enforcing geometric consistency and visual fidelity.

\paragraph{Novel View Synthesis and Video-to-Video Generation}
Generating unseen viewpoints from posed images or videos has advanced significantly in recent years~\cite{nerfstudio, mildenhall2021nerf, kerbl3Dgaussians, barron2022mipnerf360, yang2023emernerf, Yu2024GOF, Li_2023_CVPR, adaptiveshells2023, Huang2DGS2024, barron2021mipnerf, mueller2022instant, luiten2023dynamic, duan:2024:4drotorgs}. Conventional novel view synthesis (NVS) frameworks leverage volumetric or Gaussian-based scene representations; meanwhile, feed-forward architectures~\cite{yu2020pixelnerf, mvsnerf, zhou2023nerflix, tang2024lgm, jin2024lvsmlargeviewsynthesis, wang2021ibrnet, pixelsplat, chen2025mvsplat, ren2024scube} aim to directly predict target views from sparse or multi-view input, but usually struggle in generalization tasks or under challenging domain shifts. Some recent works attempt to harness image/video generative models to infuse prior knowledge and regularize deficits of view synthesis, as in ReconFusion~\cite{wu2024reconfusion} and CAT3D~\cite{gao2024cat3d}. However, these strategies tend to be slow due to per-scene optimization and often depend on robust inter-view alignment, as seen in ReconX~\cite{liu2024reconx} and ViewCrafter~\cite{yu2024viewcrafter}, which become error-prone in the presence of thin or ambiguous structures. Relatedly, the video-to-video generation field~\cite{wang2018video, wang2019few, mallya2020world, rerender, videop2p, wang2024your, chen2024follow, zhou2024upscale, xu2024videogigagan} explores producing temporally consistent and controllable video outputs under various manipulation and conditioning tasks. Techniques such as GCD~\cite{vanhoorick2024gcd}, Recapture~\cite{recapture}, GS-DiT~\cite{gs-dit}, DAS~\cite{das}, and recent 4D-consistent pipelines~\cite{ren2025gen3c, yu2025trajectorycrafter, trajattn} exploit geometric and dynamic scene information, such as tracked 3D points~\cite{cotracker, spatialtracker}, to condition or align the generative models, either via simulation or real-world sequences. These approaches enable synchronizing generated outputs across multiple cameras or time but are typically constrained by how accurately dynamic scene content can be retrieved or tracked. Some works tackle 4D and multi-view video generation by training generative models directly on synchronized video collections~\cite{sv4d, wang20254real, wang20254realv2, syncammaster, cat4d, li2024vivid}, or by reconstructing an explicit scene representation first and then rendering views~\cite{monst3r, megasam, liu2024reconx, sun2024dimensionx}.

\paragraph{Consistency in Video Generation}
Ensuring frame-to-frame and cross-view temporal or geometric consistency remains a critical challenge in video synthesis. Early efforts used 3D point clouds or height maps derived from input or generated content to guide learning and enforce consistency~\cite{mallya2020world, deng2024streetscapes}. Others~\cite{kuang2024collaborative} propagate consistency across parallel generated video streams but may lose coherence when scene elements leave the views of all sequences. Latent feature histories have also been used to improve consistency for streaming or autoregressive video generation approaches~\cite{streamingt2v}, though explicit, interpretable 3D control remains an open research direction.

%% file: sec/3_method.tex
\section{Method}
\label{sec:method}
\subsection{Preliminary: Video Diffusion Models}
Our framework builds on the popular architecture used in state-of-the-art text-to-video diffusion models~\cite{cogvideox,cosmos,hunyuanvideo,wan2025wan}, which combine a 3D Variational Auto-Encoder (VAE) with a Diffusion Transformer (DiT)~\cite{dit}. The VAE spatially and temporally compresses input videos into low-resolution latents that serve as compact representations for diffusion modeling. The diffusion process follows the rectified flow formulation~\cite{liu2022flow}, where the model learns to transform noise into coherent video latents through velocity prediction.

Within this framework, the DiT operates directly in latent space. It first patchifies the 3D latent tensor into spatio-temporal tokens $\mathbf{z}_{xyt}\in\mathbb{R}^d$, where $x$, $y$, and $t$ denote the spatial and temporal coordinates of each $d$-dimensional token. These tokens are then processed through a stack of transformer blocks, each comprising a 3D spatio-temporal self-attention layer to capture motion and appearance consistency, a text cross-attention layer for semantic conditioning, and a feed-forward network (FFN) for feature transformation. This architecture enables efficient large-scale training and produces high-quality, temporally consistent video generations.

In this work, we extend the capabilities of DiT-based diffusion architectures to support a wider range of camera control and multi-view synthesis tasks.

\input{figures/fig_approach}

\subsection{Network architecture}

As illustrated in Figure~\ref{fig:approach}, the proposed model takes as input a set of images captured from different viewpoints and time steps, represented using Plücker ray maps. The objective is to denoise the target tokens to generate video frames at any user-specified viewpoint and time. The configurations of context and target viewpoints and timestamps are flexible, enabling the model to adapt to a wide range of tasks.

To realize this capability, we next investigate the optimal designs for integrating context-frame conditioning, camera control, and the corresponding training strategies that best support these functionalities.

\noindent\textbf{Context image conditioning.}
To facilitate flexibility and scalability in the number of inputs to our model, we propose a network that performs token concatenation as inputs to the DiT. The network takes as input a set of context tokens $\mathbf{z}_{xyt}^\text{ctx}$ (encoded from multiple input views), which are concatenated token-wise with a set of target tokens $\mathbf{z}_{xyt}^\text{tgt}$, where $x$, $y$, and $t$ denote the spatial and temporal coordinates of the token vector. During the flow matching process, the target tokens are progressively denoised, where $x$, $y$, and $t$ denote the spatial and temporal coordinates of the token vector. During the flow matching process, the target tokens are progressively denoised while attending to the clean context tokens, which serve as conditioning inputs to guide generation. The overall input to the DiT is therefore represented as a joint sequence composed of both $\mathbf{z}^\text{ctx}$ and $\mathbf{z}^\text{tgt}$.

\noindent\textbf{Camera embeddings.}
To incorporate camera information, we utilize Plücker ray maps $\mathbf{P} \in \mathbb{R}^{6 \times H \times W}$, which represent the camera ray direction and origin for each image pixel. Our camera encoder $\mathcal{E}_c$ divides the ray map into patch volumes with the same spatio-temporal compression rate as the video VAE + DiT patchifier. These patch volumes are flattened channel-wise and passed through an MLP to obtain a set of camera tokens for both context and target frames, denoted as $\mathbf{c}_{xyt}^\text{ctx}$ and $\mathbf{c}_{xyt}^\text{tgt}$, with resolution and channels matching that of the video tokens. Separate camera encoders are employed for each DiT layer, allowing the model to flexibly modulate the influence of camera conditions at different stages of the network.

A naive strategy for injecting camera tokens $\mathbf{c}_{xyt}$ is to simply concatenate or add them to the corresponding video tokens $\mathbf{z}_{xyt}$. A similar approach is adopted in \cite{bai2025recammastercameracontrolledgenerativerendering}, where a camera encoder produces a 12-dimensional pose embedding that is added to the video tokens in every DiT block. However, this formulation entangles the spatial location of the camera and the temporal position of the corresponding frame within the video, as discussed below.

\noindent\textbf{Disentangle camera and temporal position embeddings.}
Video DiTs use 3D Rotary Positional Embeddings (RoPE)~\cite{su2024roformer} to encode the spatial-temporal positions $(x, y, t)$ of video tokens. 
Specifically, RoPE applies a frequency-based rotation on the keys and values of a token:
\begin{equation}
\text{RoPE}(\mathbf{q}_{xyt}^z) = R(\mathbf{q}_{xyt}^z, \thetaxyt),
\label{eq:rope}
\end{equation}
where $\thetaxyt$ denotes the sinusoidal phase parameters set by the position $(x,y,t)$, see supplementary for details. $\mathbf{q}_{xyt}^z$ denotes the query vector corresponding to the video tokens $\mathbf{z}$, and Eq.~\ref{eq:rope} is similarly applied to the key vectorss $\mathbf{k}_{xyt}^z$.

When camera embeddings $\mathbf{c}_{xyt}$ are directly added to the video tokens prior to applying 3D RoPE, the transformation becomes:
\begin{equation}
R(\mathbf{q}_{xyt}^z + \mathbf{c}_{xyt}^z, \thetaxyt) = R(\mathbf{q}_{xyt}^z, \thetaxyt) + R(\mathbf{c}_{xyt}, \thetaxyt),
\end{equation}
since RoPE is a linear projection.
This formulation, however, entangles camera and temporal information: the camera embeddings are rotated according to their specific camera corresponding timestampss $t$, even though they should ideally remain temporally invariant. Consequently, the model tends to overfit to the specific camera or unseen camera trajectories seen during training, as it implicitly encodes temporal correlations into the camera embeddings. This reduces generalization to novel or unseen camera trajectories, as further discussed in \cref{ssec:ablaton}.

To address this issue and disentangle camera and temporal representations, we propose the following approach:

\noindent\textbf{(i) Setting $t$ as a constant for camera tokens.} To eliminate the temporal interference on the camera tokens, we propose fixing their temporal index to a constant value, i.e., $t = 0$, effectively reducing the 3D RoPE to a 2D form. Under this modification, separate RoPE transformations are applied to the video tokens and the camera tokens as follows:
\begin{equation}
    \tilde{\mathbf{q}^z}_{xyt} = R(\mathbf{q}_{xyt}, \thetaxyt), \quad \tilde{\mathbf{c}}_{xyt} = R(\mathbf{c}_{xyt}, \boldsymbol{\theta}_{xy0}).
\end{equation}
For brevity, the corresponding formulation for the key vectors $\tilde{\mathbf{k}}^{z}_{xyt}$ of the video tokens is omitted.
% where $\tilde{\mathbf{q}^z}$, $\tilde{\mathbf{z}}$ denotes query vectors and camera tokens after applying RoPEs.

%for token locations $m = (x, y, t)$. This results in camera embeddings independent of the video timestamp index.
\noindent\textbf{(ii) Channel-wise concatenation of video and camera tokens.}
The RoPE-transformed camera tokens $\tilde{\mathbf{c}}$ and the video queries and keys $\tilde{\mathbf{q}}^z$, $\tilde{\mathbf{k}}^z$ must be fused before being fed into the scaled dot-product attention. This fusion can be performed either additively or via channel-wise concatenation. To analyze the design choices, we compare the resulting attention scores under both strategies. 

Let $m = (x, y, t)$ and $n = (x', y', t')$ denote the spatial-temporal positions of the query and key tokens, respectively. Under additive fusion, the attention score is given by:
\begin{equation}
    A_{n,m}^{\text{add}} = \langle \tilde{\mathbf{q}^z}_m + \tilde{\mathbf{c}}_m, \, \tilde{\mathbf{k}^z}_n + \tilde{\mathbf{c}}_n \rangle,
\end{equation}
where $\langle \cdot, \cdot \rangle$ denotes the inner product.
We observe that $A_{n,m}^{\text{add}}$ introduces entangled cross-terms between the camera and temporal positions, \ie $\langle \tilde{\mathbf{q}}^z_m, \tilde{\mathbf{c}}_n \rangle$ and $\langle \tilde{\mathbf{k}}^z_n, \tilde{\mathbf{c}}_m \rangle$, which can lead to undesirable deviations in the attention map and unstable interactions between spatial-temporal and camera embeddings.

Instead, we advocate for \emph{channel-wise concatenation} of the video and camera tokens, yielding an attention score of:
\begin{equation}
    A_{n,m}^{\text{cat}} = \langle [\tilde{\mathbf{q}^z}_m ; \tilde{\mathbf{c}}_m], \, [\tilde{\mathbf{k}^z}_n ; \tilde{\mathbf{c}}_n] \rangle = \langle \tilde{\mathbf{q}^z}_m , \, \tilde{\mathbf{k}^z}_n \rangle+\langle \tilde{\mathbf{c}}_m , \, \tilde{\mathbf{c}}_n \rangle,
\end{equation}
where the camera tokens and video tokens are fully disentangled. In the canonical case where two frames share the same camera configuration, regardless of their temporal index, the concatenation formulation yields a constant offset in the attention map, thereby maintaining behavior closely aligned with the original temporal attention structure, as desired. We show ablations by exploring the two RoPE variants in Table~\ref{tab:ablation_camera_rope} supporting our hypothesis.

\noindent\textbf{(iii) Separate QK projections for camera tokens.} Finally, we find it crucial to further enhance the representation capacity of camera tokens by introducing independent query and key (QK) projection layers that transform the camera embeddings $\mathbf{c}$ into $\mathbf{q}^c$ and $\mathbf{k}^c$. This modification allows the model to learn camera-specific attention patterns distinct from those of the video tokens.

The resulting transformer architecture, illustrated in Fig.~\ref{fig:approach}, computes the attention score as:
\begin{equation}
    A_{n,m}^{\text{cat}} = \langle \tilde{\mathbf{q}^z}_m , \, \tilde{\mathbf{k}^z}_n \rangle+\langle \tilde{\mathbf{q}^c}_m , \, \tilde{\mathbf{k}^c}_n \rangle,
\end{equation}
where $\tilde{\mathbf{q}^c}_m$, $\tilde{\mathbf{k}^c}_n$ denote the 2D RoPE-transformed query and key vectors for the camera tokens.

% Let $\langle \cdot, \cdot \rangle$ denote the inner product between two tokens at positions $m=(x,y,t)$ and $n=(x',y',t')$. 
% The two fusion variants produce attention scores given by 
% \begin{align}
% A_{n,m}^{\text{add}} &= \langle \tilde{z}_m + \tilde{c}_m, \, \tilde{z}_n + \tilde{c}_n \rangle, \\
% A_{n,m}^{\text{cat}} &= \langle [\tilde{z}_m ; \tilde{c}_m], \, [\tilde{z}_n ; \tilde{c}_n] \rangle = \langle \tilde{z}_m , \, \tilde{z}_n \rangle+\langle \tilde{c}_m , \, \tilde{c}_n \rangle,
% \end{align}
% We observe that the additive variant introduces further entanglement between the video and camera tokens, leading to substantial deviations in the resulting attention maps. In contrast, the concatenation variant acts as a consistent bias to the original attention, preserving the relative relationships between video tokens. In the canonical case where two frames share the same camera configuration, regardless of their temporal index, the concatenation formulation yields a constant offset in the attention map, thereby maintaining behavior closely aligned with the original temporal attention structure, as desired. We show ablations by exploring the two RoPE variants in \sharath{[]} supporting our hypothesis.

\subsection{Training setup}
\label{ssec:method_training_setup}
\input{tables/tab_datasets_config}

To enable the model to accommodate different input configurations across a variety of tasks, we leverage a diverse collection of existing 3D/4D datasets encompassing multiple data types. Stereo4D~\cite{jin2025stereo4d} comprises stereo videos with camera-pose annotations; however, because per-video stereo baselines are unavailable, we use a single view with its corresponding pose, effectively treating it as a monocular video dataset. We include RE10K~\cite{zhou2018re10k} and DL3DV~\cite{ling2024dl3dv} as multi-view image NVS datasets. For image/video NVS, we additionally use the synthetic Syncammaster~\cite{syncammaster} and Recammaster~\cite{bai2025recammastercameracontrolledgenerativerendering} datasets, featuring multiple time-synchronized static cameras and dynamic camera trajectories, respectively. An overview of the datasets and the tasks each dataset supports is provided in~\cref{tab:dataset_config}. We primarily target context and target configurations with $1$, $3$, $5$, or $10$ latent frames, and up to $3$ views for the context input. Owing to the variation in input views and frames across training iterations, we demonstrate in \cref{fig:n3dv_views} that our model generalizes to longer video sequences and a greater number of views. Moreover, despite not training directly on the multi-view video NVS task (e.g., $3{\times}3 \rightarrow 1{\times}3$), the model generalizes to it naturally, as it can be viewed as a composition of the multi-view image NVS and monocular video NVS tasks.

During training, we randomly sample a task and then select a dataset that supports it. Static and dynamic NVS tasks sample identical timestamps for context and target frames. In contrast, for camera-control (CamCtrl) tasks, we sample target frames such that the first target index $t_0^\text{tgt}$ lies within a temporal offset $\Delta$ (a task-dependent hyperparameter) of the first context index $t_0^\text{ctx}$, \ie, $|t_0^\text{tgt} - t_0^\text{ctx}| \le \Delta$. This enables I2V and V2V not only for consecutive future timestamps relative to the inputs but also for earlier frames where the context precedes the target, thereby enhancing temporal versatility with last-frame/middle-frame image conditioning.

%% file: figures/fig_approach.tex
\begin{figure*}[t]
  \centering
  \includegraphics[width=\linewidth]{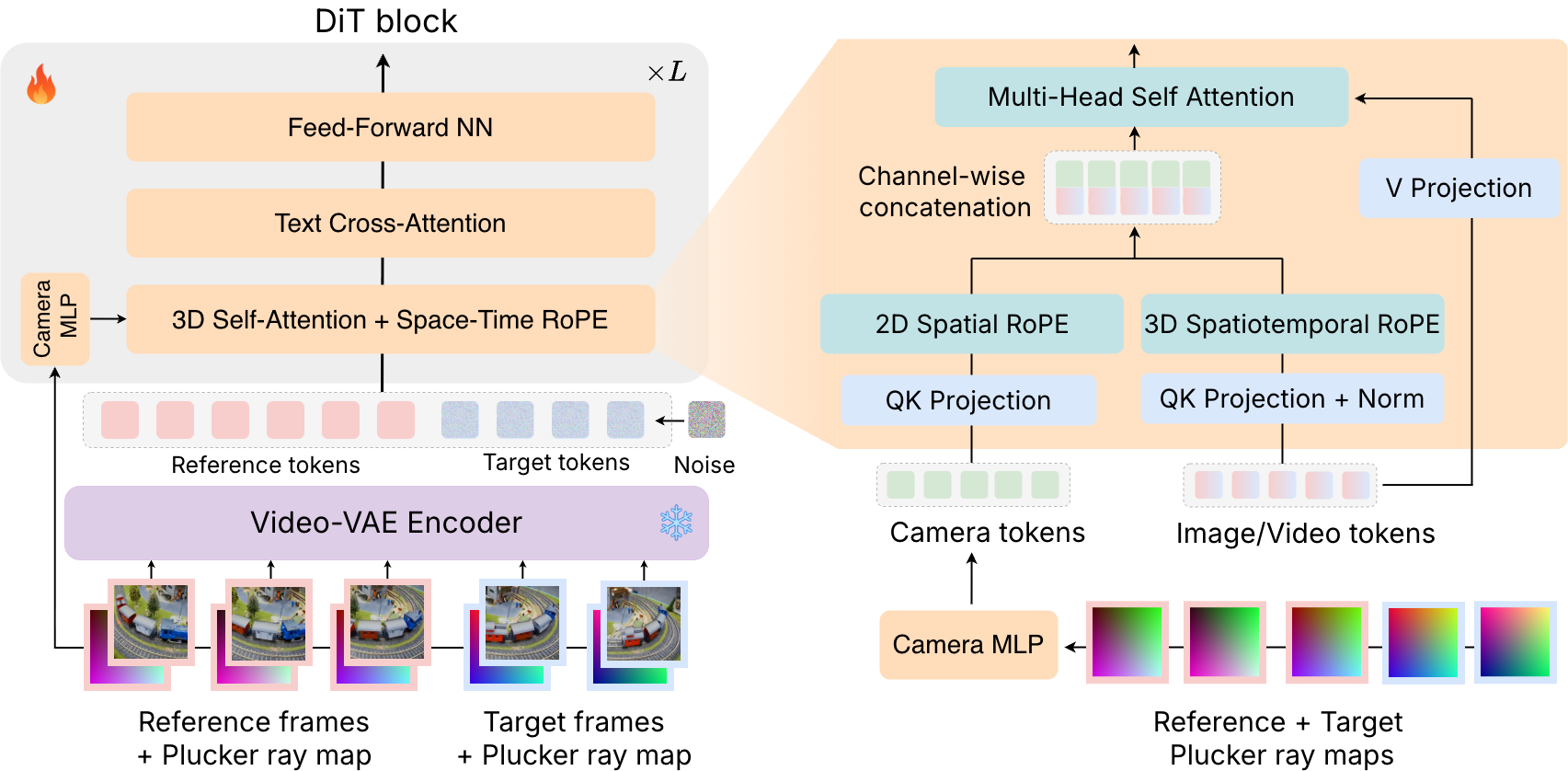}
  \caption{Overview of the network architecture. We concatenate source input tokens and partly denoised target tokens as input to the DiT. Right: We apply an MLP to the camera embeddings for each view followed by separate RoPE mechanisms for camera and video tokens. The two sets of tokens are channel-wise concatenated when input to the Self-Attention block.}
  \label{fig:approach}
  \vspace{-6pt}
\end{figure*}

%% file: tables/tab_datasets_config.tex
\begin{table}
\small
\setlength{\tabcolsep}{0pt}
\caption{Multitask training configurations for various datasets in terms of number of views V and number of latent frames F.}
\label{tab:multitask-config}
\begin{tabularx}{\linewidth}{lZYY}
\toprule
\textbf{Task Type} & \textbf{Datasets} & \makecell{\textbf{Context}\\ $(V \times F)$} & \makecell{\textbf{Target } \\ $(V \times F)$}\\
\midrule
\makecell[l]{Monocular\\Image NVS, \\ Multi-view\\Image NVS } & \makecell{RE10K,\\DL3DV,\\ReCamMaster,\\ SynCamMaster}&
\makecell{3$\times$1\\2$\times$1\\1$\times$1} &
\makecell{1$\times$1} \\
\cmidrule{1-4}
\makecell[l]{Monocular\\Video NVS}& \makecell{ReCamMaster, \\SynCamMaster}&
\makecell{1$\times$3} &
\makecell{1$\times$3} \\
\cmidrule{1-4}
T2V + CamCtrl & \makecell{RE10K, \\DL3DV, \\Stereo4D} &
 - &
\makecell{1$\times$3\\1$\times$5\\1$\times$10} \\

\cmidrule{1-4}
I2V + CamCtrl& \makecell{RE10K, \\DL3DV, \\Stereo4D} &
\makecell{1$\times$1} &
\makecell{1$\times$3\\ 1$\times$5\\ 1$\times$10}\\
\cmidrule{1-4}
V2V + CamCtrl& \makecell{RE10K, \\DL3DV, \\Stereo4D} &
\makecell{1$\times$3} &
\makecell{1$\times$3\\ 1$\times$5\\ 1$\times$10} \\
\bottomrule
\end{tabularx}
\label{tab:dataset_config}
\end{table}

%% file: sec/4_experiments.tex
\section{Experiments}
\label{sec:experiments}

\input{figures/fig_qualitative_davis}
The key takeaways of our experiments are: a) \method is capable of high quality 4D consistency tasks across a wide variety of settings, including camera control and novel view synthesis for both static and dynamic scenes. b) Compared to specialized methods that focuses on specific settings, \method effectively combines many types of camera and time conditions via our proposed RoPE, leading to generalization across a variety of tasks. c) As the number of input views increase, \method is able to effectively leverage the increased signal in the input to improve reconstruction quality. d) Through ablations, we show that our proposed camera RoPE design is crucial for effective modeling of camera and time conditions, leading to improved performance across tasks.

\subsection{Experimental Setup}
\label{ssec:exp_setup}

\textbf{Implementation Details.} As discussed in~\cref{ssec:method_training_setup}, we train \method on a mixture of the datasets provided in~\cref{tab:dataset_config}. We choose the Wan 2.1 1.1B model~\cite{wan2025wan} as the base architecture for all our experiments unless mentioned otherwise. Each iteration takes in a set of source views, target views and their corresponding cameras as well as their timestamps in the real world. We train the model for 40K iterations over 32 H100 GPUs. We linearly warmup the learning rate for 3K iterations upto $0.0001$ with a batch size of 64. During the warmup stage, we exclusively train on the multiview static task to quickly update the parameters corresponding to the Plucker ray maps while also allowing the model to quickly adapt to camera conditioning. 

\input{tables/tab_quantitative_davis}
\textbf{Evaluation and Baselines.} We comprehensively evaluate \method on a wide variety of 3D/4D tasks. We broadly group multiple tasks under: a) Monocular Video NVS (\cref{tab:quantitative_davis}) b) Multi-view Image and Video NVS (\cref{tab:quantitative_llff}) and c) T2V/I2V + Camera Control. We mainly perform comparisons with closely related generative view-synthesis works of RecamMaster~\cite{bai2025recammastercameracontrolledgenerativerendering}, TrajectoryCrafter~\cite{yu2025trajectorycrafter}, and GEN3C~\cite{ren2025gen3c} on the Monocular Video NVS as well as the I2V + Camera Control task. For Multi-view Image NVS, GEN3C and SEVA~\cite{zhou2025stable} are our primary baselines which can take multiple view inputs.  We visualize the results from our model in \cref{fig:qualitative_davis} and comparisons in the supplementary. We use the reconstruction metrics of PSNR, SSIM, LPIPS for datasets with GT views available. We additionally evaluate the quality of the camera trajectory via Rotation (RotErr) and Translation Error (TrErr) as proposed by~\cite{cameractrl}. We use MegaSaM~\cite{megasam} for estimating camera trajectories from generated videos when comparing with the GT trajectories. We additionally compute the metric scale for each scene similar to~\cite{ac3d} and scale the camera translation vectors accordingly. We find this necessary for monocular view cases which have inherent scale ambiguity. 

\subsection{Monocular Video NVS}
\label{ssec:mono_video_nvs}

For this task, we extract camera trajectories for 45 real-world videos from DAVIS~\cite{Huang_2016_davis}. For each video, we use 4 predefined trajectory paths from RecamMaster evaluations and 1 new trajectory which is a spiral path. Results are summarized in~\cref{tab:quantitative_davis}. Averaged over all trajectories, we match the performance of SOTA approaches of~\cite{bai2025recammastercameracontrolledgenerativerendering,yu2025trajectorycrafter,ren2025gen3c} in terms of camera error metrics despite being trained on a large diversity of tasks. We additionally visualize qualitative results in the supplementary. We see that we obtain high fidelity generations compared to other works with fewer motion artifacts. We maintain high temporal synchronization compared to the input condition highlighting the capability of the model to accurately extract appearance information from conditioning views while disentangling camera and time conditions.

Beyond camera trajectory error metrics, we also compare with~\cite{yu2025trajectorycrafter,ren2025gen3c} on the Neural 3D Video (N3DV) dataset~\cite{li2022neural} which consists of multi-view static cameras capturing a dynamic scene. We choose the first view as the test and choose one training view with significant disparity from the test. Compared to prior works, we obtain high quality reconstructions which are well aligned with the GT achieving better PSNR, SSIM and LPIPS. Note that our conditioning is largely implicit via encoded latents and we do not use any form of explicit 3D supervision such as depth or point clouds apart from the metric scene scale. Despite being trained on 122K camera trajectories, RecamMaster  fails for this setting of static target camera  and tends to implicitly induce camera motion. This highlights the generalizability issue of using 3D spatiotemporal RoPE on camera conditions as discussed in~\cref{sec:method} where the model fails to disentangle camera and time resulting in poor out-of-distribution trajectory performance.

\subsubsection{Multi-view Static and Dynamic NVS}
\label{ssec:multi_image_video_nvs}
\input{tables/tab_quantitative_llff}
\input{figures/fig_qualitative_n3dv}

We now extend our approach to the multi-view input setting consisting of images or time-synchronized videos. We first evaluate \method on the LLFF dataset~\cite{mildenhall2019locallff} consisting of multiple view captures of a static scene. We sparsely sample 3, 6, or 9 input views and choose one test view. We compare against SEVA~\cite{zhou2025stable} and GEN3C as they allow for multiple view inputs. Results are summarized in~\cref{tab:quantitative_llff}, measuring the 3 reconstruction metrics. We see that we perform competitively against the baselines across 3 metrics. While GEN3C is limited by its cache size which is 4 by default, we directly generalize to more input views despite being trained on upto 3 static views, while improving reconstruction quality with additional 3D information of the scene.

This is observed in the dynamic setting as well for N3DV, where we continue to improve reconstruction quality from 1 to 5 views~\cref{fig:n3dv_views} in terms of PSNR, SSIM, and LPIPS. We visualize the reconstructions in~\cref{fig:qualitative_n3dv}. As the number of views increase, the model is able to better resolve the scale ambiguity with static cameras producing increasingly more aligned generations with the GT view.

Notably, our training configuration (\cref{tab:dataset_config}) does not include the multi-view dynamic NVS task but only on multi-view static NVS and monocular dynamic NVS. This highlights the capability of our model to generalize to not only more views or more frames, but also to new 3D/4D tasks which are combinations of trained tasks. This opens the avenue to potentially include a variety of inputs such as multiple view combinations of images and videos.
\input{figures/fig_n3dv_views}

\subsection{T2V/I2V + Camera Control}
\input{tables/tab_quantitative_re10k}
\input{tables/tab_t2v_i2v_camerror.tex}
While NVS targets generations with timestamps which are same as input conditions, T2V/I2V + Camera Control involves target timestamps which are in the future from the image condition for I2V. Additionally, we show that the model also generalizes to T2V by passing no conditional views but only specifying target trajectories. We evaluate both T2V/I2V on RE10K on a subset of 1000/2000 samples respectively. We compare with TrajectoryCrafter and Gen3C on I2V, and with AC3D on T2V. Results are summarized in~\cref{tab:quantitative_re10k_i2v,tab:quantitative_re10k_t2v}.

We see that we obtain better reconstructions aligned with the GT views compared to~\cite{yu2025trajectorycrafter,ren2025gen3c} while also more closely aligning with the camera trajectory compared to AC3D~\cite{ac3d}. This is despite AC3D being trained primarily on the Camera Control task while we train on a variety of different 3D/4D tasks. Qualitative results are provided in the supplementary to show high video fidelity obtained even in the case of no input conditioning views.

% \subsection{Novel View Synthesis for Static Scenes}

% \subsection{Camera Control for Dynamic Scenes}

% \subsection{Camera Control for Image-to-Video}

% \subsection{Camera Control for Text-to-Video}

% \subsection{Combining Signals}

% \subsection{Out of Distribution Camera Trajectories?}

% \subsection{Generalization Compared to Baselines}

\subsection{Ablation Study: Camera RoPE}
\label{ssec:ablaton}

% \subsubsection{Camera RoPE Variants}

\input{tables/tab_ablation_camera_rope}

We conduct an ablation study to analyze the contribution of different components in our camera RoPE approach. Table~\ref{tab:ablation_camera_rope} shows the performance of various configurations on the N3DV dataset.

Our final approach with attention head concatenation achieves the best performance across all metrics, demonstrating the effectiveness of our design choices.

%% file: figures/fig_qualitative_davis.tex
\begin{figure*}[t]
  \centering
  \includegraphics[width=\linewidth]{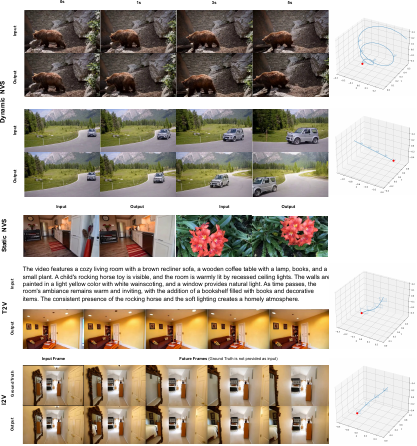}
  \caption{\textbf{Example generations from \method.} We show results on (a) dynamic NVS; (b) static NVS; (c) text-to-video (T2V) with camera control; and (d) image-to-video (I2V) with camera control. In all cases, \method produces high-fidelity, 3D-consistent videos that adhere to the input conditioning(s).}
  \label{fig:qualitative_davis}
\end{figure*}

%% file: tables/tab_quantitative_davis.tex
\begin{table*}
    [t]
    \centering
    \caption{Quantitative comparison on different camera trajectory types on
    DAVIS. We perform competitively against prior SOTA Video NVS approaches while outperforming in certain categories.}
    \label{tab:quantitative_davis} \small
    \renewcommand{\arraystretch}{1.2}
    \setlength{\tabcolsep}{4pt}
    \begin{tabularx}
        {\textwidth}{l | *{2}{>{\centering\arraybackslash}X} *{2}{>{\centering\arraybackslash}X} *{2}{>{\centering\arraybackslash}X} *{2}{>{\centering\arraybackslash}X} *{2}{>{\centering\arraybackslash}X} | *{2}{>{\centering\arraybackslash}X} }
        \toprule \multirow{2}{*}{Method} & \multicolumn{2}{c}{Trans Up} & \multicolumn{2}{c}{Trans
        Down} & \multicolumn{2}{c}{Arc Left} & \multicolumn{2}{c}{Arc Right} & \multicolumn{2}{c}{Spiral
        w/ Zoom} & \multicolumn{2}{|c}{Overall} \\
        \cmidrule(lr){2-3} \cmidrule(lr){4-5} \cmidrule(lr){6-7} \cmidrule(lr){8-9}
        \cmidrule(lr){10-11} \cmidrule(lr){12-13} & TrErr$\downarrow$ & RotErr$\downarrow$
        & TrErr$\downarrow$ & RotErr$\downarrow$ & TrErr$\downarrow$ & RotErr$\downarrow$
        & TrErr$\downarrow$ & RotErr$\downarrow$ & TrErr$\downarrow$ & RotErr$\downarrow$
        & TrErr$\downarrow$ & RotErr$\downarrow$ \\

        \midrule
        TrajectoryCrafter~\cite{yu2025trajectorycrafterredirectingcameratrajectory}
        & 24.77 & 3.59 & 21.36 & 3.60 & 14.37 & 4.00 & 24.92 & 2.64 & 23.11 & 2.05 & 21.70 & 3.18 \\
        ReCamMaster~\cite{bai2025recammastercameracontrolledgenerativerendering}
        & \textbf{10.18} & 2.60 & \textbf{11.47} & \underline{2.19} & \underline{8.85} & 3.39 & \underline{16.49} & 2.82
        & \textbf{13.49} & 2.37 & \textbf{12.09} & 2.67\\
        Gen3C~\cite{ren2025gen3c}
        & 28.38 & \textbf{1.88} & 42.55 & \textbf{2.04} & 129.17 & \textbf{2.62} & 21.80 & \textbf{1.39}
        & 25.72 & \textbf{1.71} & 49.52 & \textbf{1.93} \\
        \midrule
        \textbf{\method} 
        & \underline{11.75} & \underline{2.20} & \underline{20.51} & 2.46 & \textbf{5.77} & \underline{3.27} & \textbf{8.88} & \underline{2.50}
        & \underline{14.53} & \underline{2.03} & \underline{12.29} & \underline{2.49} \\
        \bottomrule

        % TrajectoryCrafter~\cite{yu2025trajectorycrafterredirectingcameratrajectory}
        % & 24.77 & 3.59 & 21.36 & 3.60 & 14.37 & 4.00 & 24.92 & 2.64 & 23.11 & 2.05 & 21.70 & 3.18 \\
        % \midrule ReCamMaster~\cite{bai2025recammastercameracontrolledgenerativerendering}
        % & 10.18 & 2.60 & 11.47 & 2.19 & 8.85 & 3.39 & 16.49 & 2.82 & 13.49 & 2.37
        % & 12.09 & 2.67\\
        % Gen3C~\cite{ren2025gen3c}
        % & 28.38 & 1.88 & 42.55 & 2.04 & 129.17 & 2.62 & 21.80 & 1.39 & 25.72 & 1.71 & 49.52 & 1.93 \\
        % \midrule
        % \textbf{\method} & 11.75 & 2.20 & 20.51 & 2.46 & 5.77 & 3.27 & 8.88 & 2.50
        % & 14.53 & 2.03 & 12.29 & 2.49 \\
        % \bottomrule
    \end{tabularx}
\end{table*}

%% file: tables/tab_quantitative_llff.tex
\begin{table}
    [t]
    \centering
    \caption{Quantitative comparison on LLFF dataset. We outperform Stable Virtual Camera~\cite{zhou2025stable} significantly in terms of reconstruction quality for the static NVS task.}
    \label{tab:quantitative_llff} \small
    \renewcommand{\arraystretch}{1.2}
    \setlength{\tabcolsep}{6pt}
    \begin{tabular}{l | c c c c}
        \toprule Method & Views &PSNR$\uparrow$ & SSIM$\uparrow$ & LPIPS$\downarrow$ \\
        \midrule
        SEVA~\cite{zhou2025stable} & \multirow{2}{*}{3} &14.84&0.30&0.46 \\
        \textbf{\method} &&\textbf{15.43}&\textbf{0.38}&\textbf{0.41} \\
        \midrule
        SEVA~\cite{zhou2025stable} & \multirow{2}{*}{6} &15.36&0.32&0.43 \\
        \textbf{\method} &&\textbf{16.11}& \textbf{0.42}& \textbf{0.38} \\
        \midrule
        SEVA~\cite{zhou2025stable} & \multirow{2}{*}{9} &15.60&0.33&0.42 \\
        \textbf{\method} &&\textbf{16.49}& \textbf{0.45}& \textbf{0.34} \\
        \bottomrule
    \end{tabular}
\end{table}

%% file: figures/fig_qualitative_n3dv.tex
\begin{figure}[t]
  \centering
  \includegraphics[width=\linewidth]{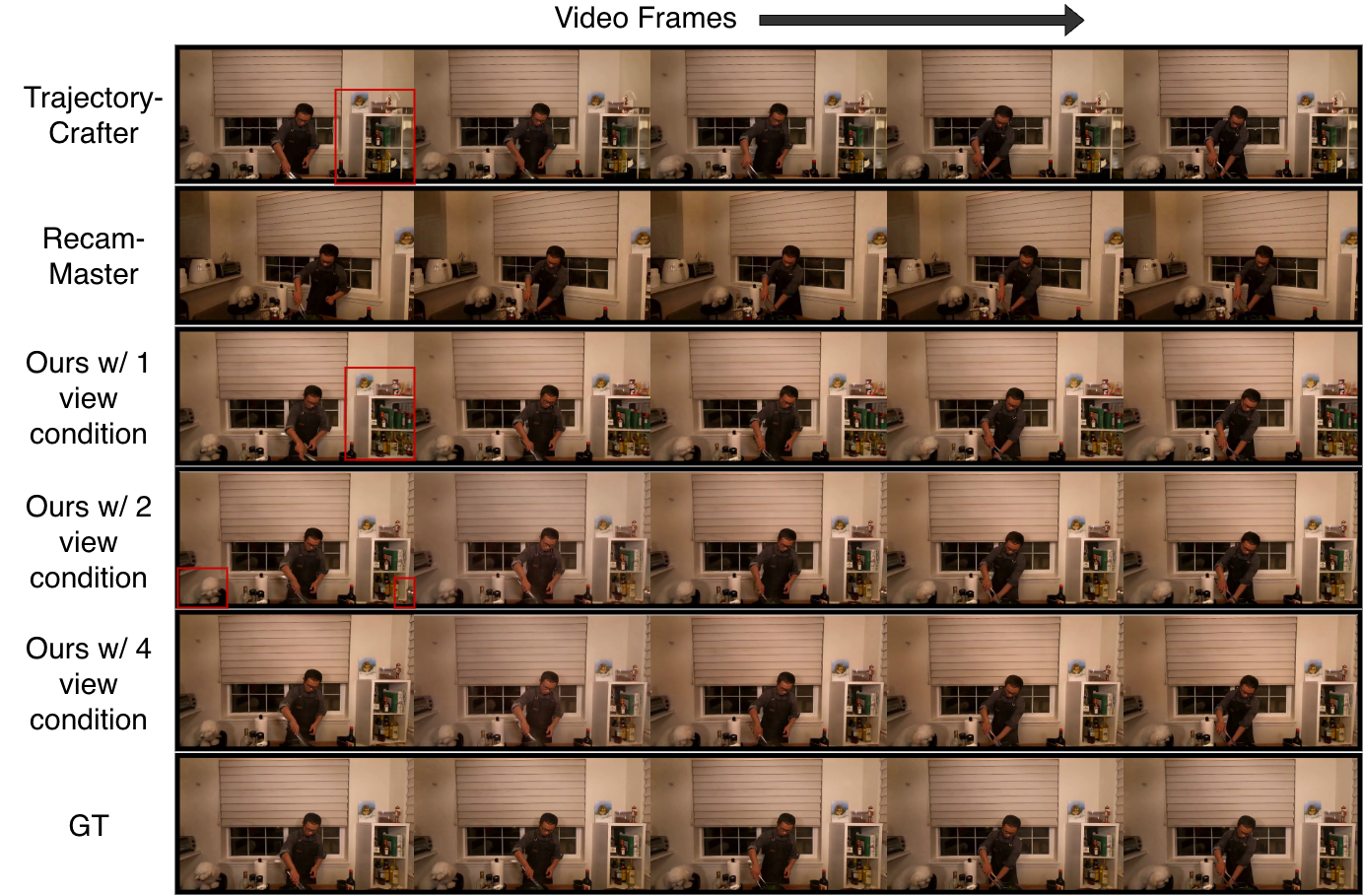}
  
  % \setlength{\fboxsep}{0pt}
  % \fbox{\rule{0pt}{0.22\textheight}\rule{\linewidth}{0pt}}
  \caption{Qualitative visualizations on a scene in the N3DV dataset. We consistently outperform \cite{bai2025recammastercameracontrolledgenerativerendering,yu2025trajectorycrafter} on single view input while improving reconstruction quality with increasing number of input conditions.}
  \label{fig:qualitative_n3dv}
  \vspace{-6pt}
\end{figure}

%% file: figures/fig_n3dv_views.tex
\begin{figure}[t]
  \centering
  \includegraphics[width=\linewidth]{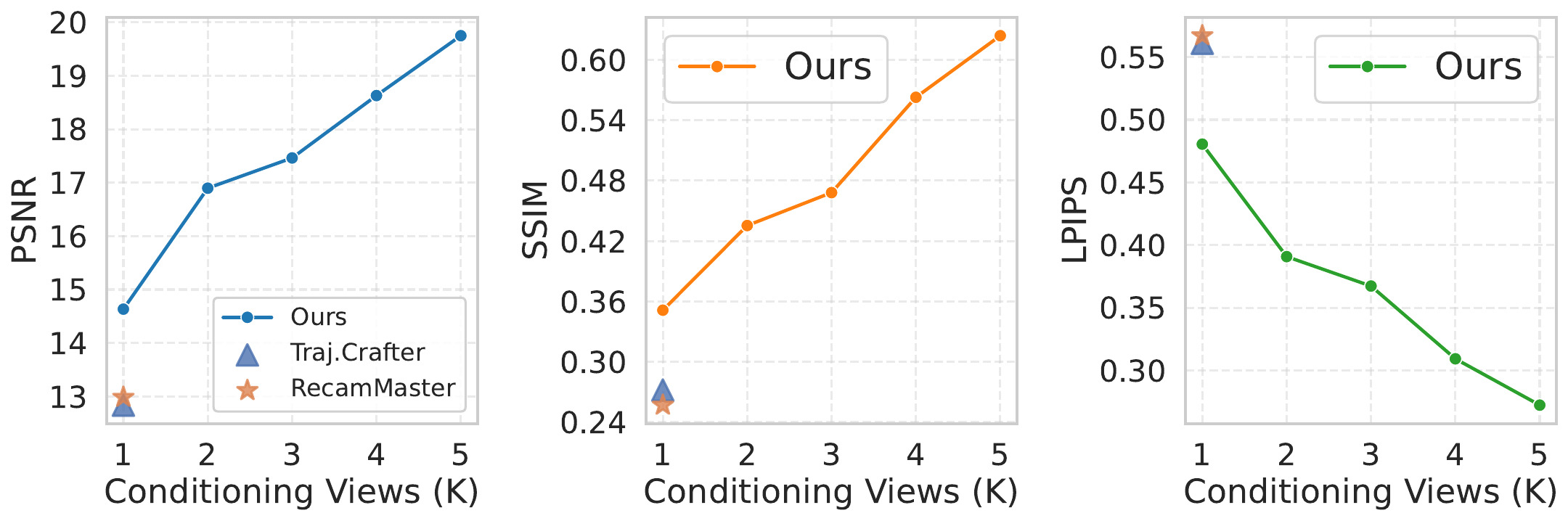}
  \caption{Increasing number of conditional views improves reconstruction on N3DV~\cite{li2022neural} and is more aligned with the GT.}
  \label{fig:n3dv_views}
  \vspace{-6pt}
\end{figure}

%% file: tables/tab_quantitative_re10k.tex
\begin{table}
    [t]
    \centering
    \caption{Quantitative comparison on RE10K dataset. We continue to outperform~\cite{yu2025trajectorycrafter,ren2025gen3c} on the Image-to-Video + Camera Control task across all reconstruction metrics.}
    \label{tab:quantitative_re10k_i2v} \small
    \renewcommand{\arraystretch}{1.2}
    \setlength{\tabcolsep}{6pt}
    \begin{tabular}{l | c c c}
        \toprule Method & PSNR$\uparrow$ & SSIM$\uparrow$ & LPIPS$\downarrow$ \\
        \midrule
        TrajectoryCrafter~\cite{yu2025trajectorycrafterredirectingcameratrajectory} & 16.94 & 0.53 & 0.36 \\
        Gen3C~\cite{ren2025gen3c} & 17.34 & 0.55 & 0.34 \\
        \midrule
        \textbf{\method} & \textbf{19.20} & \textbf{0.66} & \textbf{0.28} \\
        \bottomrule
    \end{tabular}
    \vspace*{0em}
\end{table}

%% file: tables/tab_t2v_i2v_camerror.tex
\begin{table}
    [t]
    \centering
    \caption{Camera-error metrics on RE-10K for the T2V+Camera Control task. We outperform AC3D across both metrics. }
    \label{tab:quantitative_re10k_t2v} \small
    \renewcommand{\arraystretch}{1.2}
    \setlength{\tabcolsep}{4pt}
    \begin{tabular}{l | c c}
    \toprule
    \textbf{Metric} & AC3D~\cite{ac3d} & \textbf{\method} \\
    \midrule
    TransErr$\downarrow$ & 5.170 & \textbf{1.412} \\
    RotErr$\downarrow$   & 1.365 & \textbf{0.572} \\
    \bottomrule
    \end{tabular}
\end{table}

%% file: tables/tab_ablation_camera_rope.tex
\begin{table}
    [t]
    \centering
    \caption{Ablation study exploring different RoPE variants.}
    \label{tab:ablation_camera_rope} \small
    \renewcommand{\arraystretch}{1.2}
    \setlength{\tabcolsep}{6pt}
    \begin{tabular}{l | c c c}
        \toprule Variant & PSNR$\uparrow$ & SSIM$\uparrow$ & LPIPS$\downarrow$ \\
        \midrule
        No RoPE applied to Plucker & 13.36 & 0.292 & 0.554 \\
        Apply 3D RoPE to Plucker & 13.68 & 0.309 & 0.509 \\
        Apply 2D RoPE to Plucker & 14.17 & 0.345 & 0.504 \\
        \ \ + Plucker to values & 14.25 & 0.334 & 0.524 \\
        \midrule
        \textbf{Ours: + Attn Cat} & \textbf{15.46} & \textbf{0.376} & \textbf{0.456} \\
        \bottomrule
    \end{tabular}
\end{table}

%% file: sec/5_conclusion.tex
\section{Conclusion}
\label{sec:conclusion}
In this work, we introduced \method, a novel framework for 4D consistent video generation that effectively integrates camera and time conditions using our camera RoPE. Our approach demonstrates superior performance across various tasks, including camera control and novel view synthesis for both static and dynamic scenes. Through extensive experiments and ablation studies, we validated the effectiveness of our design choices, particularly the use of RoPE for modeling camera and time conditions. Future work could explore further enhancements to the model architecture and investigate additional applications in 4D content generation.

%% file: sec/X_suppl.tex
\clearpage
\setcounter{page}{1}
\maketitlesupplementary

\section{Background: RoPE}
\label{supp:rope_preface}
% Rotary Positional Embeddings (RoPE)~\cite{su2024roformer} provides functionality for incorporating positional information into tokens for transformer self-attention blocks
RoPE~\cite{su2024roformer} injects positional information into self-attention by \emph{rotating} each 2D channel pair of the query and key vectors with sinusoidal, index-dependent phases. This rotation makes attention scores a function of relative displacement (via phase differences) rather than absolute indices, yielding a smooth decay with distance, translation equivariance of the dot-product, and strong length/extrapolation behavior—all without adding parameters. For video tokens, the position is naturally 3D $(x,y,t)$; phases can be assigned per axis and combined (e.g., additively) to induce a spatio–temporal inductive bias. We use the shorthand
\(\text{RoPE}(\mathbf{q}_{xyt}^z)=R(\mathbf{q}_{xyt}^z,\thetaxyt)\)
and apply the same operator to keys, where \(\thetaxyt\) denotes the sinusoidal phase parameters determined by $(x,y,t)$.

Let the token dimension be $d$ and index $J=d/2$ two-dimensional channel pairs by $j=0,\dots,J{-}1$.
For each axis $a\in\{x,y,t\}$ define
\[
\omega^{(a)}_j \;=\; B^{-\,2j/d},\qquad B=10{,}000.
\]
For a video position $p=(x,y,t)$, the per-pair phase is
\[
\varphi_j(p)\;=\;x\,\omega^{(x)}_j \;+\; y\,\omega^{(y)}_j \;+\; t\,\omega^{(t)}_j,
\qquad
\thetaxyt \equiv \{\varphi_j(p)\}_{j=0}^{J-1}.
\]
Splitting a vector $u\in\mathbb{R}^{d}$ into 2D slices $u^{(j)}=(u_{2j},u_{2j+1})^\top$ and rotating each slice gives us
\[
R(u,\thetaxyt)^{(j)} \;=\;
\begin{bmatrix}\cos\varphi_j & -\sin\varphi_j\\ \sin\varphi_j & \cos\varphi_j\end{bmatrix}
u^{(j)},\quad \varphi_j=\varphi_j(p).
\]
This is then applied to queries and keys from video latents $\mathbf{z}$:
\[
\hat{\mathbf{q}}_{xyt}^z=R(\mathbf{q}_{xyt}^z,\thetaxyt),\qquad
\hat{\mathbf{k}}_{xyt}^z=R(\mathbf{k}_{xyt}^z,\thetaxyt).
\]
For positions $m$ and $n$, the inner product depends only on the relative phase:
\[
\langle \hat{\mathbf{q}}_{m}^z,\, \hat{\mathbf{k}}_{n}^z \rangle
\;=\;
\langle \mathbf{q}_{m}^z,\; R(\mathbf{k}_{n}^z,\,\{\varphi_j(n)-\varphi_j(m)\}_j)\rangle,
\]
yielding a smooth relative spatio-temporal bias without added parameters.

\section{Ablation with PRoPE}
In addition to the ablations shown with different variants in Table 6 of the main paper, we compare with another variation of RoPE for camera control~\cite{li2025camerasrelativepositionalencoding} (PRoPE). PRoPE consists of replacing the RoPE rotation matrix with the camera projection matrices for half of the dimensions while reorganizing the other half for 2D xy coordinates. They, however, require training the diffusion model from initialization as opposed to our approach which utilizes existing trained video diffusion models. Nevertheless, we evalute their approach in a finetuning setting using the pretrained video diffusion model. As they do not directly target the dynamic video setting, we extend their approach to modify the RoPE dimensions corresponding to the spatial coordinates while keeping the temporal dimensions as is. We train the model in this setting for 30K iterations and evaluate on the N3DV monocular video NVS task. Results are summarized in~\cref{supptab:ablation_prope}. We see that PRoPE leads to noisy reconstructions across all metrics likely due to the model incapable of significantly altering the underlying RoPE mechanism in the DiT, and can potentially require a large number of iterations for convergence. This is in contrast to our approach, which closely retains the 3D spatio-temporal RoPE structure, especially in the standard base setting of same cameras between different latents.

\input{figures/supp_fig_qualitative_llff_compare}
\input{figures/supp_fig_llff_views}
\input{figures/supp_fig_re10k}
\input{figures/supp_fig_re10k_t2v}
\input{figures/supp_fig_davis}

\section{Qualitative visualizations}
We now visualize the generations from our approach across a wide-variety of tasks

\paragraph{Static multi-view NVS on LLFF} We visualize the generations from our approach as well as Stable Virtual Camera (SEVA)~\cite{zhou2025stable}  for the case of 3 conditioning views and compare it with the Ground-Truth (GT) across 5 scenes in LLFF. Results are shown in~\cref{supp_fig:qualitative_llff_compare}. We see that we consistently obtain higher quality generations while SEVA leads to blurrier outputs. We also obtain generations which are more aligned with the GT such as the tip of the fern (top row), or the flower (second from top row), or the legs of the dinosaur (bottom row). We also visualize the generations with increasing number of views in~\cref{supp_fig:llff_views}. With increasing multi-views as input, the model reduces scale ambiguity reconstructing the scene more faithfully while also incorporating appearance information from additional input views.

\input{tables/supp_tab_ablation_prope}
\paragraph{T2V/I2V+Camera Control on RE10K} 
We show qualitative results with Camera Control on RE10K with text (T2V) or image (I2V) conditions. I2V results are visualized in~\cref{supp_fig:qualitative_re10k}. We see that compared to GEN3C or TrajectoryCrafter we more closely align with the input camera trajectory resulting in generations which are better aligned with the GT view. We highlight certain regions with red boxes such as the door in the first column, or the photo frame in the last column showing generations with reference to the GT. We also produce realistic hallucinations in new regions unseen in the input view, while largely maintaining 3D consistency with the regions visible in the input.

\noindent Next, we compare the camera control performance for the T2V setting with AC3D~\cite{ac3d}. Results are shown in~\cref{supp_fig:qualitative_re10k_t2v}. We utilize the camera trajectory from a source video and generate new videos with this trajectory. We see that AC3D can signficantly deviate from the source trajectory, such as the zooming in for the first scene while AC3D rotates left. We on the hand, are consistent with the camera input equivalently zooming into the scene. A similar but inverse result is observed in the third scene where the input camera slightly pans/rotates left while AC3D  produces a zoomed in version, while we produce generations consistent with this trajectory.

\paragraph{Monocular Video NVS on DAVIS}
We now show results for the monocular video NVS task on DAVIS. Results are visualized in~\cref{supp_fig:davis} comparing with GEN3C and Recammaster. We show 3 target trajectories from top to bottom for the 3 scenes: a) Translating right while looking to the left) b) Translating down while looking up and c) Translating left while looking to the right.
The input trajectory can vary depending on the scene. We see that we consistently obtain clean generations. For the first scene, we arc right showing a slightly frontal view of the swan while GEN3C produces significant deformations and Recammaster mainly translates with little rotation. Both approaches also fail for the second scene resulting in significantly high artifacts in the generation inconsistent with the input. For the 3rd view with the target trajectory close to the input trajectory, we produce relatively clean generations with fewer artifacts close to the bus.

% \section{N3DV / LLFF / Davis, 60K (latest ckpt) results}

% \section{DAVIS / RE10K / N3DV, 14B results}

% \section{RoPE (promised)}

% \section{Qualitative for RE10K}

% \section{T\&T results}

% \section{Gen3C on N3DV}

% \section{AI videos}

% \subsection{video nvs (2/4/6 secs, spiral without increasing radius), Frame conditioning: first frame, last frame, interpolation, static nvs, first k frames}

% 1. split into 8, 16, 24 fps
% 2. split into megasam+metric rescale, megasam+metric rescale / 2, 0 poses
% 3. split into different target poses (spiral1 == increase radius fast then constant radius, spiral2 == constant radius, original at the center)
% 4. split into tasks (static nvs == dup frames for input, move cam for output, moving + freeze + moving (17, 16, 16))

%% file: figures/supp_fig_qualitative_llff_compare.tex
\begin{figure*}[t]
  \centering
  \includegraphics[width=0.75\linewidth]{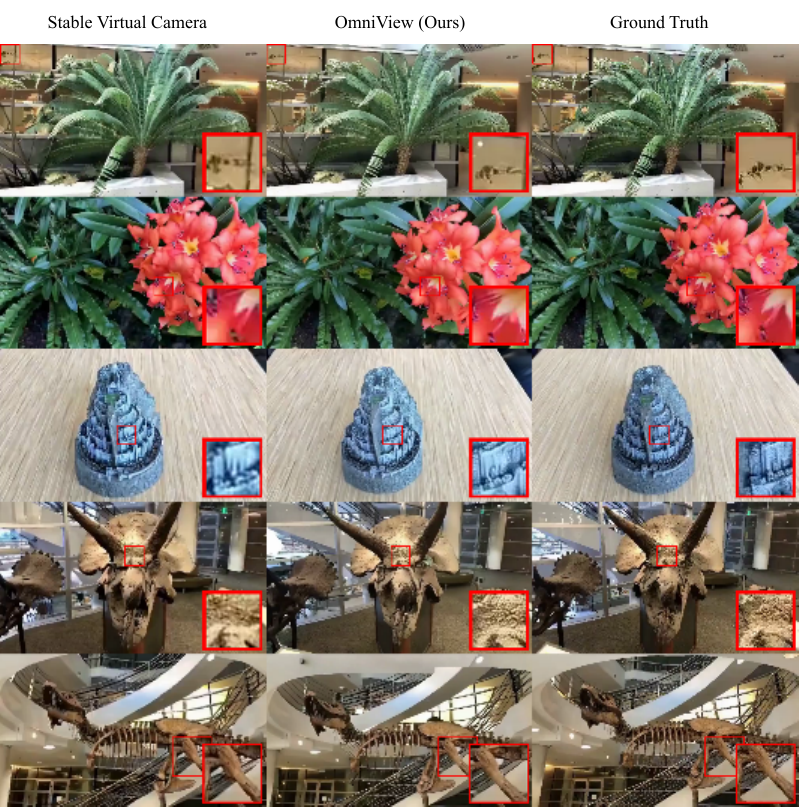}
  
  % \setlength{\fboxsep}{0pt}
  % \fbox{\rule{0pt}{0.22\textheight}\rule{\linewidth}{0pt}}
  \caption{\textbf{Qualitative visualizations on the static multi-view NVS task on the LLFF dataset} Compared to Stable Virtual Camera~\cite{zhou2025stable}, we obtain much higher detail along with aligned generations to the Ground-Truth, highlighted by the larger gap in SSIM values in Table 3 of the main paper.}
  \label{supp_fig:qualitative_llff_compare}
\end{figure*}

%% file: figures/supp_fig_llff_views.tex
\begin{figure*}[t]
  \centering
  \includegraphics[width=\linewidth]{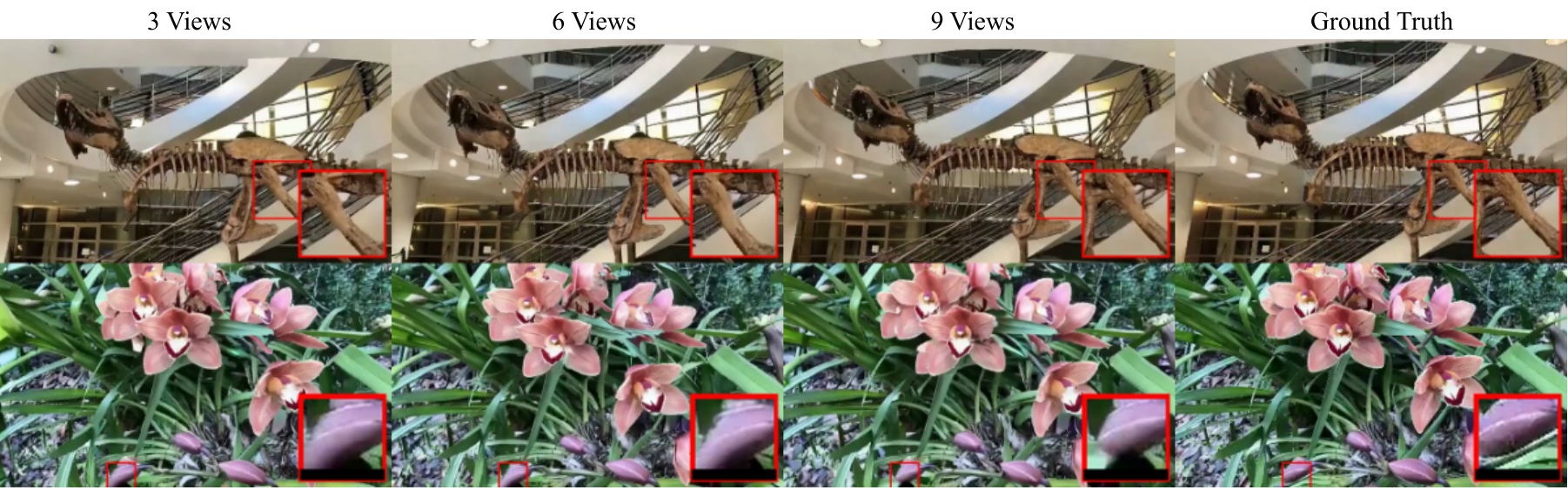}
  
  % \setlength{\fboxsep}{0pt}
  % \fbox{\rule{0pt}{0.22\textheight}\rule{\linewidth}{0pt}}
  \caption{\textbf{Varying number of input views for static multi-view NVS.} We consistently improve upon our generations with increasing number of input conditioning views leading to reconstructions more aligned with the Ground-Truth.}
  \label{supp_fig:llff_views}
\end{figure*}

%% file: figures/supp_fig_re10k.tex
\begin{figure*}[t]
  \centering
  \includegraphics[width=\linewidth]{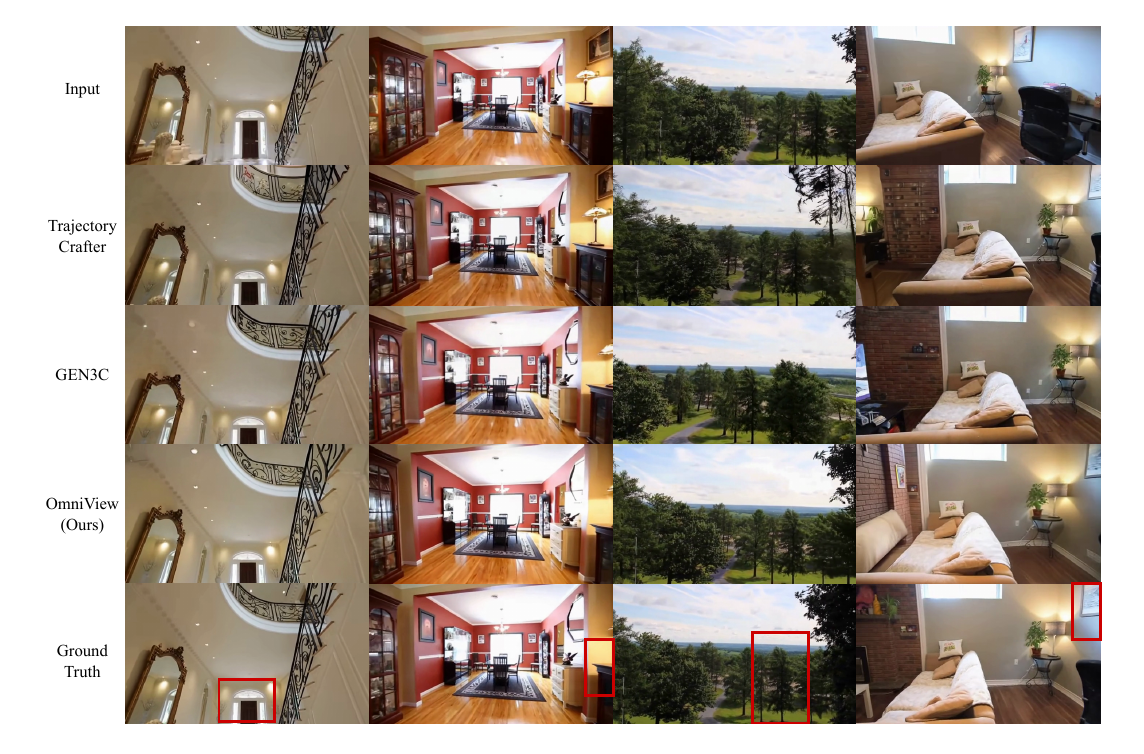}
  
  % \setlength{\fboxsep}{0pt}
  % \fbox{\rule{0pt}{0.22\textheight}\rule{\linewidth}{0pt}}
  \caption{\textbf{Qualitative visualizations on the I2V + Camera Control task on RE10K.} Compared to prior works of~\cite{yu2025trajectorycrafter,ren2025gen3c}, we obtain generations which are more 3D consistent with the input image condition and hence are better aligned with the Ground-Truth (Highlighted red boxes for example).}
  \label{supp_fig:qualitative_re10k}
\end{figure*}

%% file: figures/supp_fig_re10k_t2v.tex
\begin{figure*}[t]
  \centering
  \includegraphics[width=\linewidth]{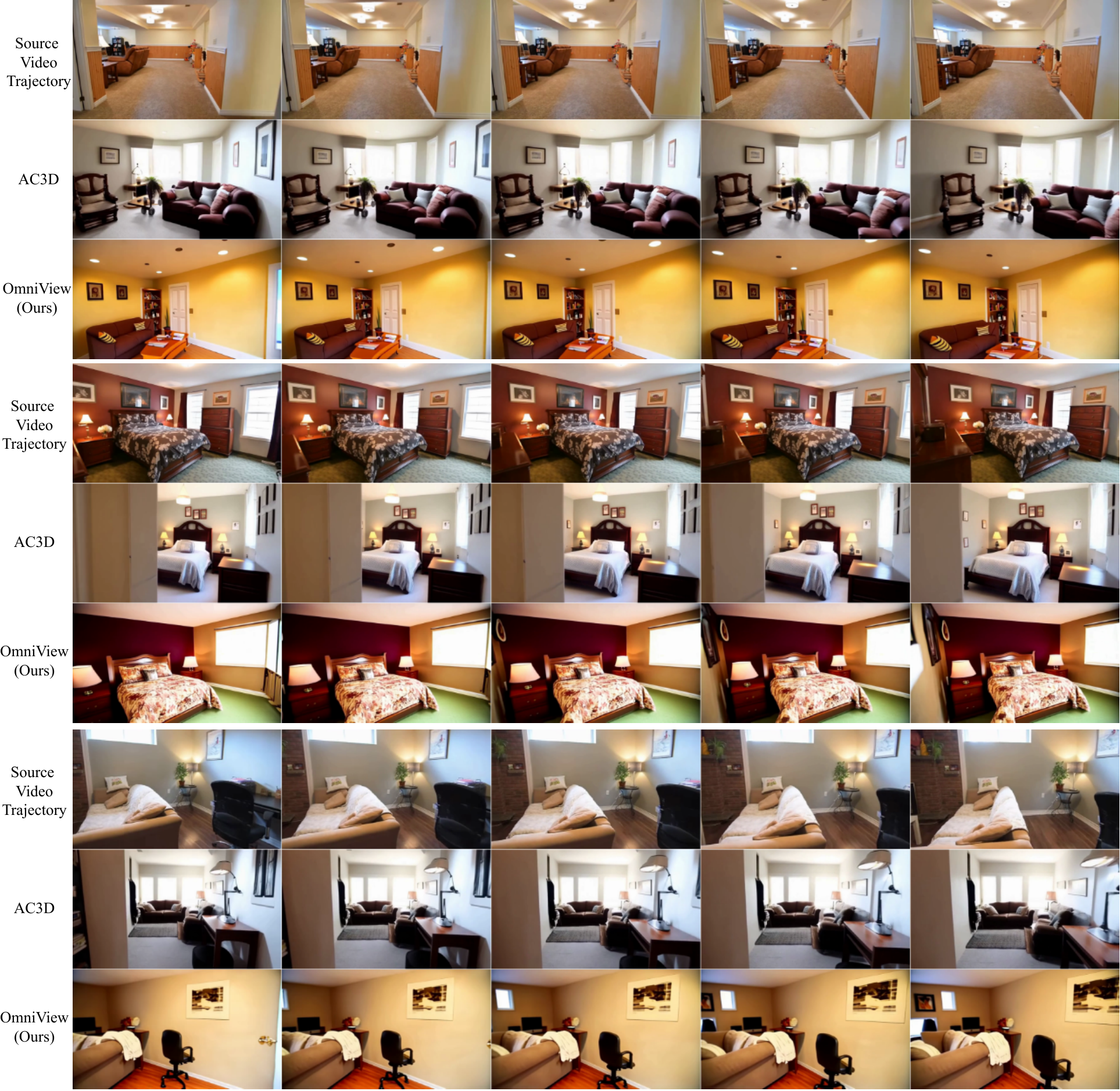}
  
  % \setlength{\fboxsep}{0pt}
  % \fbox{\rule{0pt}{0.22\textheight}\rule{\linewidth}{0pt}}
  \caption{\textbf{Qualitative visualizations on the T2V + Camera Control task on RE10K.} Compared to AC3D~\cite{ac3d}, we obtain generations which follow the camera trajectory in the source video while also producing high-fidelity video generations.}
  \label{supp_fig:qualitative_re10k_t2v}
\end{figure*}

%% file: figures/supp_fig_davis.tex
\begin{figure*}[t]
  \centering
  \includegraphics[width=0.95\linewidth]{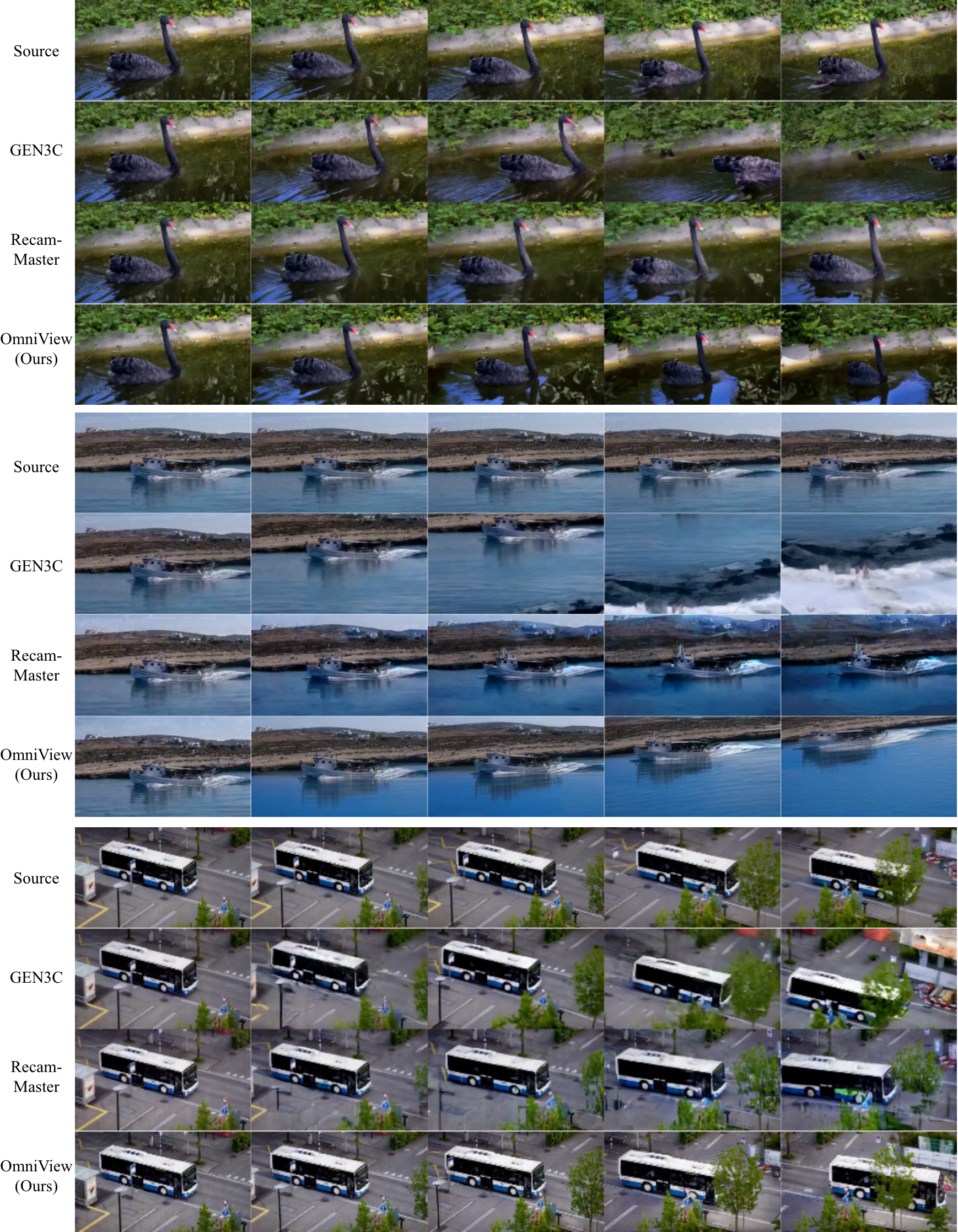}
  
  % \setlength{\fboxsep}{0pt}
  % \fbox{\rule{0pt}{0.22\textheight}\rule{\linewidth}{0pt}}
  \caption{\textbf{Video NVS generations on DAVIS.} We visualize generated videos with new trajectories compared to the source video. We obtain generations which are more trajectory consistent (top, middle), and also higher fidelity generations (bottom) with fewer artifacts.}
  \label{supp_fig:davis}
\end{figure*}

%% file: tables/supp_tab_ablation_prope.tex
\begin{table}
    [t]
    \centering
    \caption{Ablation study comparing with PRoPE~\cite{li2025camerasrelativepositionalencoding} on N3DV.}
    \label{supptab:ablation_prope} \small
    \renewcommand{\arraystretch}{1.2}
    \setlength{\tabcolsep}{6pt}
    \begin{tabular}{l | c c c}
        \toprule Variant & PSNR$\uparrow$ & SSIM$\uparrow$ & LPIPS$\downarrow$ \\
        \midrule
        PRoPE & 12.39 & 0.358 & 0.648 \\
        \textbf{Ours: + Attn Cat} & \textbf{15.46} & \textbf{0.376} & \textbf{0.456} \\
        \bottomrule
    \end{tabular}
\end{table}